\documentclass[letterpaper, 10 pt, conference]{ieeeconf}
\IEEEoverridecommandlockouts
\usepackage[utf8]{inputenc}
\usepackage[noend]{algpseudocode}
\usepackage{amsmath}
\usepackage{amssymb}
\usepackage[utf8]{inputenc}
\usepackage{booktabs}
\usepackage{multicol}
\usepackage{multirow}
\usepackage{xcolor}
\usepackage{graphicx}
\usepackage{hyperref}
\usepackage[linesnumbered,boxed,ruled,commentsnumbered]{algorithm2e}
\graphicspath{{figures/}}
\DeclareGraphicsExtensions{.pdf,.png,.jpg,.eps}

\usepackage[noadjust]{cite} 

\begin{document}
\title{\LARGE \bf STD: Stable Triangle Descriptor for 3D place recognition}
\author{Chongjian Yuan$^{12*}$,
        Jiarong Lin$^{1*}$,
        Zuhao Zou$^{1}$,
        Xiaoping Hong$^{2+}$,
        and Fu Zhang$^{1+}$,% <-this % stops a space
\thanks{$^*$These two authors contribute equally to this work.}
\thanks{$^+$Corresponding authors.}
\thanks{$^{1}$C. Yuan, J. Lin, Z. Zou and F. Zhang are with the Department of Mechanical Engineering, The University of Hong Kong, Hong Kong Special Administrative Region, People's Republic of China.
{\tt\footnotesize $\{$ycj1, zivlin, zuhao.zou$\}$@connect.hku.hk}, {\tt\footnotesize $ $fuzhang$ $@hku.hk}}
\thanks{$^{2}$C. Yuan and X. Hong are with the School of System Design and Intelligent Manufacturing, Southern University of Science and Technology, Shenzhen, People’s Republic of China.{\tt\footnotesize $\{$yuancj2020,hongxp$\}$@sustech.edu.cn}}
}

\markboth{Journal of \LaTeX\ Class Files,~Vol.~6, No.~1, January~2007}%
{Shell \MakeLowercase{\textit{et al.}}: Bare Demo of IEEEtran.cls for Journals}

\maketitle
\thispagestyle{empty}

\begin{abstract}
In this work, we present a novel global descriptor termed \textit{stable triangle descriptor (STD)} for 3D place recognition. For a triangle, its shape is uniquely determined by the length of the sides or included angles. Moreover, the shape of triangles is completely invariant to rigid transformations. Based on this property, we first design an algorithm to efficiently extract local key points from the 3D point cloud and encode these key points into triangular descriptors. Then, place recognition is achieved by matching the side lengths (and some other information) of the descriptors between point clouds. The point correspondence obtained from the descriptor matching pair can be further used in geometric verification, which greatly improves the accuracy of place recognition. In our experiments, we extensively compare our proposed system against other state-of-the-art systems (i.e., M2DP, Scan Context) on public datasets (i.e., KITTI, NCLT, and Complex-Urban) and our self-collected dataset (with a non-repetitive scanning solid-state LiDAR). All the quantitative results show that STD has stronger adaptability and a great improvement in precision over its counterparts. To share our findings and make contributions to the community, we open source our code on our GitHub: \href{https://github.com/hku-mars/STD}{\tt github.com/hku-mars/STD}.

%Experiments on the benchmark dataset (KITTI, NCLT, etc.) shows the high performance of our method compared to other state-of-the-art. Experiments on multiple types of scenarios (urban road, park and indoor) with non-repetitive scanning solid-state LiDAR further verify the adaptability of our method to different environments and LiDAR.
\end{abstract}

\IEEEpeerreviewmaketitle

\section{Introduction}
% Place recognition on a 3D point cloud refers to the problem of detecting if two 3D point clouds (e.g., a collection of points measured by range sensors) are measured for the same scene. It is a fundamental problem in many robotic techniques and applications that have increasingly been using ranging sensors (e.g., LiDARs, laser scanners)

Place recognition refers to the problem of detecting if two measurements of the sensor (e.g., camera image, LiDAR point cloud) are collected in the same scene.
It is a fundamental problem in a variety of robotic applications, such as loop detection in simultaneous localization and mapping (SLAM)\cite{lin2020loam, appearance-loop2013tro, lin2019fast}, global re-localization in prior maps \cite{1-year2019ral,rgb-relocalization}, and maps merging in multi-robot systems \cite{decentralize2020iros}.
Many vision-based SLAM systems \cite{orbslam, qin2018vins, r2live, r3live,zhu2020camvox} have been proposed due to the widespread use of cameras. However, these loop detection methods are difficult to deal with the strong variation caused by illumination, appearance, or viewpoint changes. On the other hand, Light Detection and Ranging sensor (LiDAR) is invariant to illumination and appearance change as it can directly obtain the structural information of the environment. The emergence of low-cost and high-performance LiDARs has further increased the use of LiDAR in robotics \cite{yang2022lidarVelocity, yuanpixel, fastlio2, r3live_pp}.

Generally, an efficient LiDAR-based place recognition solution should satisfy the following requirements. First, the solution is required to achieve both rotation and translation invariance regardless of the viewpoint changes. Second, the solution should better provide a relative pose. A good initial pose estimation allows the subsequent registration algorithm to converge faster and more accurate. Third, the method should be robust to different LiDAR point cloud densities and environments since the sparsity of LiDAR point cloud varies with distance, scene, and LiDAR types. 

In order to achieve the above performance, in this paper, we develop a new descriptor termed {\it stable triangle descriptor (STD)}, which encodes any three key points in the scene by a triangle. Compared with polygons used in other descriptors, the triangle is more stable because the shape of the triangle is uniquely defined given the length of the sides (or included angles). Compared with local descriptors around key points, the shape of a triangle is completely rotation and translation invariant. To extract key points for triangle descriptors, we perform point cloud projection on the plane  and extract key points on the boundary, then form the key points into triangles. 
Matches are made based on the similarity of triangles. A typical place recognition case with STD is shown in Fig. \ref{fig:descriptor_match}, which successfully recognizes two point clouds collected at opposite view angles in the same place.  Specifically, our contributions are as follows:
\begin{figure}[t]
    \centering
    \setlength{\abovecaptionskip}{-0.1cm}
    \includegraphics[width=1\linewidth]{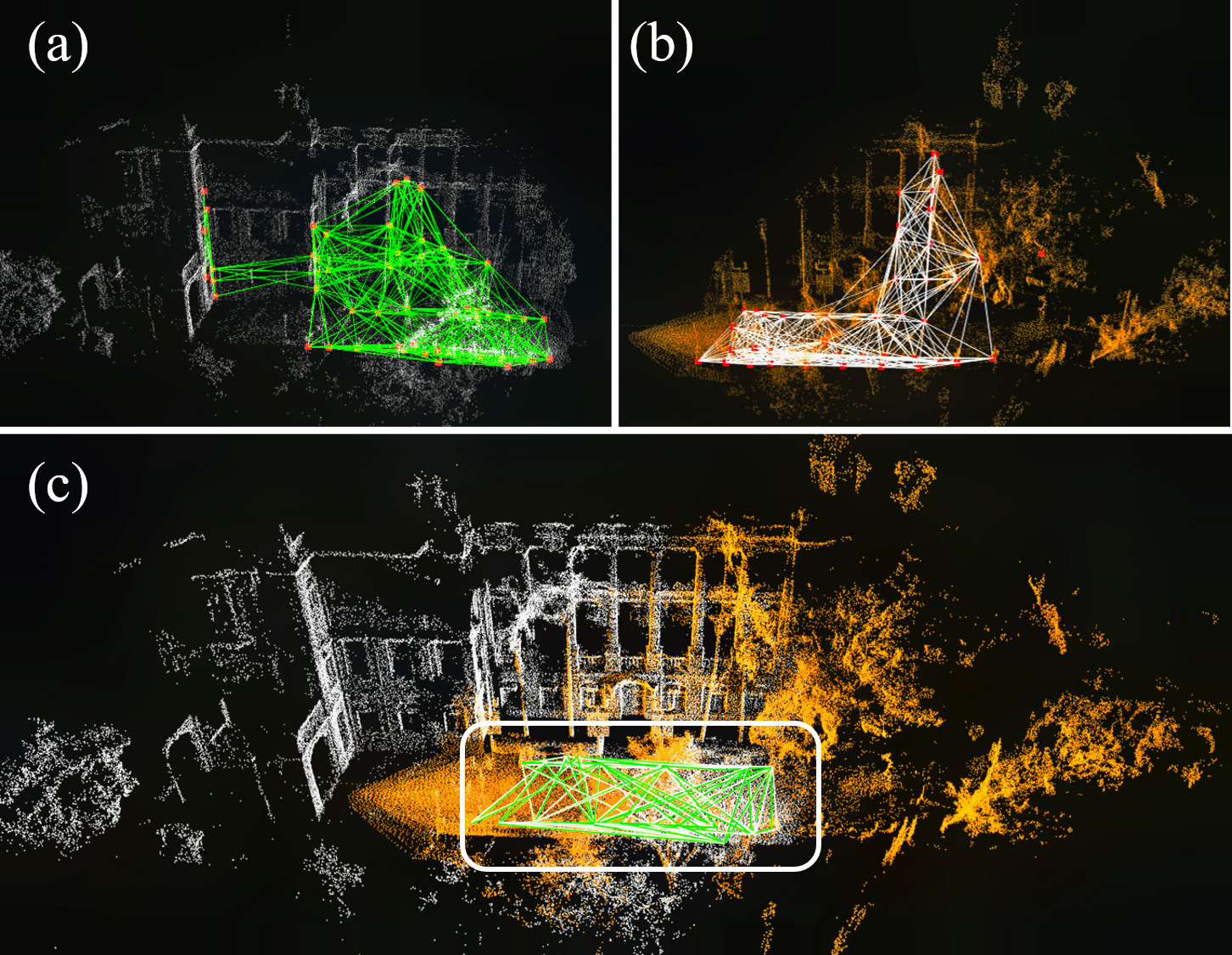}
    \caption{(a) shows the stable triangle descriptor (STD) extracted from a query point cloud. (b) shows the STD extracted from a historical point cloud. In (c),  an example of STD matching between these two frames of the point cloud. The correctly matched STD descriptors are indicated by white box, and the point clouds are registered by the poses provided by the STD. These two frames of point clouds are collected by a small FOV LiDAR (Livox Avia) moving in opposite directions, resulting in a low point cloud overlap and drastic viewpoint change (See our accompanying video on YouTube: \href{https://www.youtube.com/watch?v=O-9iXn1ME3g}{\tt youtu.be/O-9iXn1ME3g}). }
    \vspace{-0.4cm}
    \label{fig:descriptor_match}
\end{figure}
\begin{itemize}
    \item
    We design a triangle descriptor (see Fig. \ref{fig:triangle_descriptor}), a six-dimension vector consisting of the length of three triangle sides and the angles between the normal vectors of the adjacent plane attached to each triangle vertex. The descriptor is completely invariant to rotation and translation while maintaining a high degree of distinguishability. %Due to the uniqueness and stability of the triangle, the descriptor has strong rotation and translation invariance. 
    \item 
    We propose a fast key point extraction method based on keyframes. In order to represent the structural information of the scene, we project the point cloud at the plane boundary and extract key points therein, which will form triangle descriptors with adjacent key points.
    \item
    We evaluate our algorithm under multiple types of scenarios (urban, indoor, and unstructured environments) and different LiDAR data (conventional spinning LiDARs and solid state LiDARs). Sufficient experimental results verify the effectiveness of our method.
    \item To share our findings and make contributions to the community, let our readers can quickly reproduce our work in their follow-up research, we make our codes publicly available on our GitHub: \href{https://github.com/hku-mars/STD}{\tt github.com/hku-mars/STD}.
\end{itemize}

\section{Related Works}
Place recognition in 3D data is a key problem for robot localization and has been addressed in different approaches. According to the principle of the methods, we have divided the existing work into the following three categories: (i) local descriptor based on point features. (ii) global descriptor based on appearance (iii) learning-based method. 

Inspired by the place recognition solution in robotic vision, Bastian\cite{pointFeature2010icra} transforms a given 3D scan data into a range image, then extracts point features from the range image and performs score matching for 3D scan data based on point features. Bosse and Zlot \cite{keypointBosse2013icra} extract key points directly on 3D data, then use the Gestalt Descriptor that encodes each key point's neighborhood and calculate the voting matrix for place recognition. Apart from
the Gestalt descriptors, some other descriptors, such as PFH \cite{PFH}, SURFs \cite{SURF}, or SHOT \cite{SHOT}, are also used in a similar framework. However, these local descriptors are sensitive to the density and noise of the LiDAR point clouds and are not invariant to rotation or translation of the viewpoint. 

Global descriptors prefer to use the appearance information of the point cloud. Magnusson et al. \cite{ndt} first divide the point cloud into overlapping grids, then compute the shape properties of each cell by normal distribution transform, and finally combines them into a matrix of surface shape histograms. He {\it et al}. propose M2DP \cite{m2dp}, which
is generated by projecting a 3D point cloud to multiple 2D
planes and generating a high dimensional compact global representation. Giseop Kim and Ayoung Kim \cite{kim2018scan} propose scan Context, a 2D descriptor based on the height of the surrounding structures. V. Nardari {\it et al.} \cite{forest} propose a polygon descriptor to achieve place recognition in the forest environment. Jiang {\it et al} \cite{jiang2d} propose a triangle feature-based descriptor for 2D SLAM. In short, these global descriptors make use of the appearance information (surface flatness and orientation, height, and so on) of the scene.

In contrast with the local descriptors, global descriptors are more robust against noise and resolution changes, but they still struggle with the change of viewpoints. Therefore, some other works attempt to use deep learning for 3D place recognition tasks. SegMap \cite{segmatch2017icra} achieves place recognition by matching semantic features. 
% But SegMap does not perform well in the case of multiple dynamic objects since dynamic objects will alter the segmentation.
OverlapNet \cite{overlap2020rss} proposes a deep neural network to achieve overlap calculation and relative yaw angle estimates between pairs of 3D scans. These learning-based approaches all require a training step and use GPU acceleration.

Our proposed descriptor, {\it stable triangle descriptor}, is a global descriptor consisting of three key points. The triangle descriptors are extracted within a keyframe and describe the relative distribution of key points in the frame. Compared with other global descriptors \cite{ndt,m2dp,kim2018scan}, our descriptor has stronger rotation and translation invariance. Compared with the polygon descriptor \cite{forest,jiang2d}, our descriptor is extracted directly in 3D space, and uses the most recognizable and invariable triangle 
among polygons, while \cite{forest,jiang2d} extracts polygon descriptors in 2D space. In addition, our descriptor can provide pose estimation with full degrees of freedom, which can greatly reduce the registration time while ensuring the registration accuracy.

\section{Methodology}
\begin{figure*}
%    \centering
     \begin{minipage}{0.70\linewidth}
        \setlength{\abovecaptionskip}{-0.1cm}
		\includegraphics[width=1.0\linewidth]{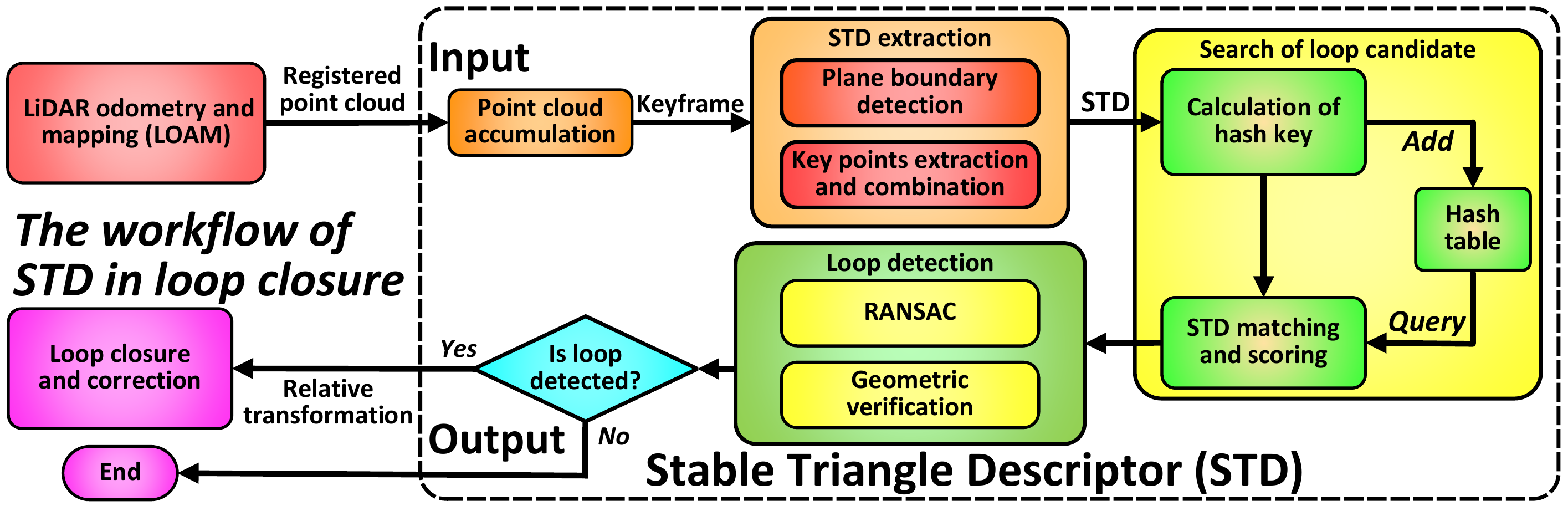}
		\caption{Workflow of our algorithm. Our method computes the triangle descriptors from keyframes. Then a hash table is used to serve as the database of our descriptors for quick store and match. Frames with the top 10 descriptor matching scores will be selected as candidates. The loop candidate is regarded as a valid loop once passing the geometric verification. The relative transformation between the loop frame and candidate frame will also be obtained as the loop is triggered.}
		\label{fig:overview}    
     \end{minipage}
      \hspace{0.05cm}
     \begin{minipage}{0.29\linewidth}
    		 \centering
    	    \includegraphics[width=1.0\linewidth]{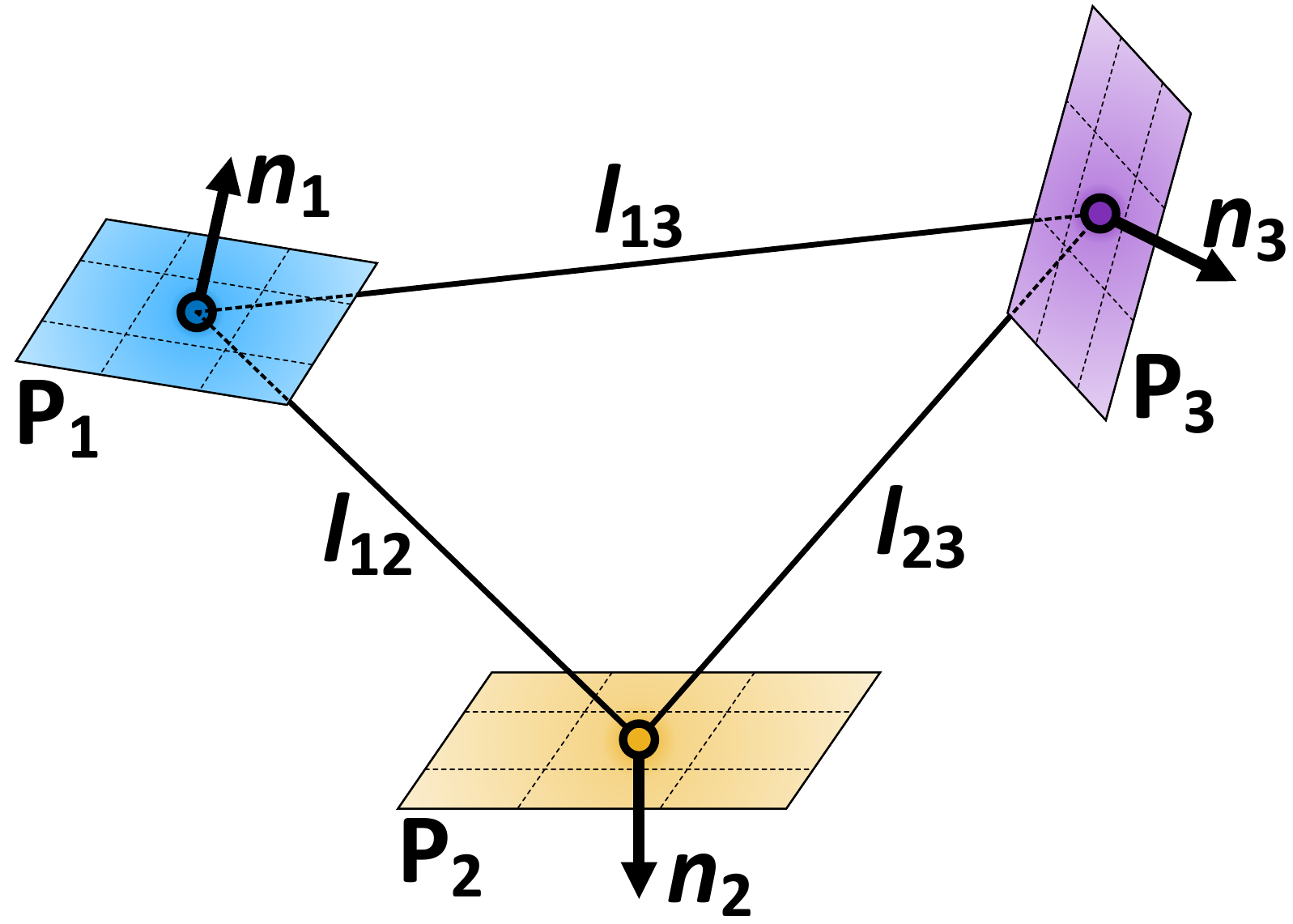}
    	    \caption{A standard triangle descriptor. Each vertex $\mathbf p_1,\mathbf p_2, \mathbf p_3$ corresponds to a adjacent plane.
    	    $\mathbf n_1, \mathbf n_2, \mathbf n_3$ are the normal vector to the 
            adjacent planes. Vertices are arranged according to $l_{12} \leq l_{23} \leq l_{13}$.}
    	    \label{fig:triangle_descriptor}
      \end{minipage}
      \vspace{-0.6cm}
\end{figure*}

In this section, we describe how to extract the stable triangle descriptor from a point cloud. Next, we introduce how to build the descriptor dictionary and how to select loop candidates. Finally, RANSAC-based loop detection and geometric verification are proposed for a complete loop detection pipeline. The overall pipeline of our method is depicted in Fig. \ref{fig:overview}.
\subsection{Stable Triangle Descriptor\label{method:descriptor}}
\begin{figure}[ht]
    \centering
    \vspace{-0.2cm}
    \setlength{\abovecaptionskip}{-0.1cm}
    \includegraphics[width=1\linewidth]{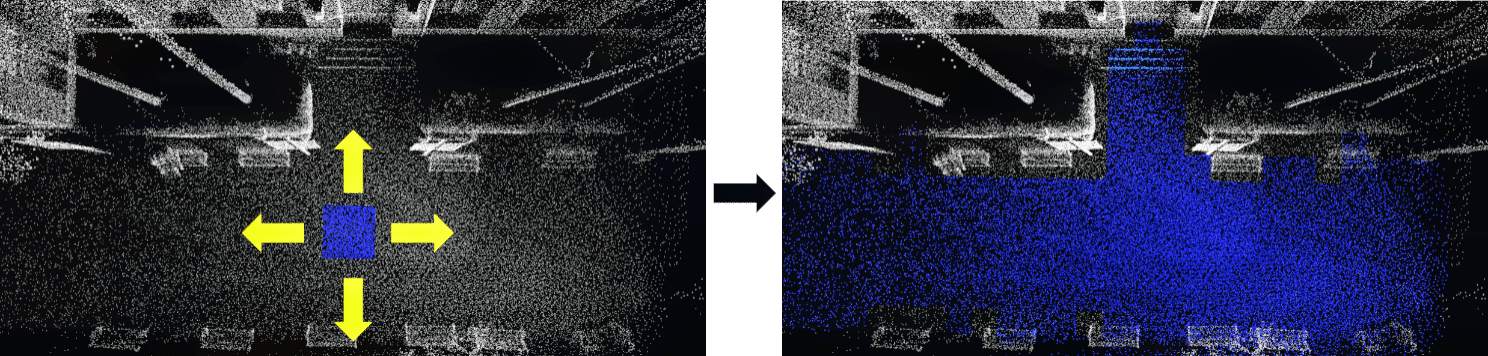}
    \caption{Plane expand process}
    \vspace{-0.4cm}
    \label{fig:plane_expand}
\end{figure}

Inspired by \cite{segmatch2017icra}, to improve the stability of segmentation, we perform loop detection on keyframes, which have points accumulated from a few consecutive scans and hence have increased point cloud density regardless of the specific LiDAR scanning pattern. Specifically, we use a LiDAR odometer \cite{yuan2022voxelMap} to register each new incoming point cloud into the current keyframe. A new keyframe will be created when the number of sub-frames accumulates to a certain number. When given a keyframe of the point cloud, we first perform plane detection by region growing. Specifically, we divide the entire point cloud into voxels of given sizes (e.g., 1m). Each voxel contains a group of points $\mathbf p_i \ (i=1,...,N)$; we then calculate the point covariance matrix $\mathbf \Sigma$:
\begin{equation}
\abovedisplayshortskip=3pt
\belowdisplayshortskip=3pt
\abovedisplayskip=3pt
\belowdisplayskip=3pt
\begin{aligned}
   \bar{\mathbf p}=\frac{1}{N}\sum_{i=1}^N \mathbf p_i; \quad \mathbf \Sigma=\frac{1}{N} \sum_{i=1}^N(  \mathbf p_i -\bar{\mathbf p})(  \mathbf p_i -\bar{\mathbf p})^T;     
\end{aligned}
\end{equation}
Let $\lambda_k$ denote the $k$-th largest eigenvalue of matrix $\mathbf \Sigma$. The plane criterion principle is: 
\begin{equation}
\abovedisplayshortskip=3pt
\belowdisplayshortskip=3pt
\abovedisplayskip=3pt
\belowdisplayskip=3pt
\begin{aligned}
    \lambda_3< \sigma_1 ~\text{and}~ \lambda_2> \sigma_2
\end{aligned}
\end{equation}
where $\sigma_1$ and $\sigma_2$ are pre-set hyperparameters. By this criterion, we can check whether the points in a voxel form a plane and if so, the voxel is called a plane voxel. Then, we initialize a plane with any plane voxel and grow the plane by searching for its neighboring voxels. If the neighboring voxels are the same planes (has the same plane normal direction with a distance below a threshold), they are added to the plane under growing. Otherwise, if the neighboring voxel is not on the same plane, it is added to a list of boundary voxels for the plane under growing. The above growing process repeats until all the added neighboring voxels are expanded, or boundary voxels are reached (see Fig. \ref{fig:plane_expand}). 

\begin{figure}[ht]
    \centering
    \vspace{-0.2cm}
    \setlength{\abovecaptionskip}{-0.1cm}
    \includegraphics[width=1\linewidth]{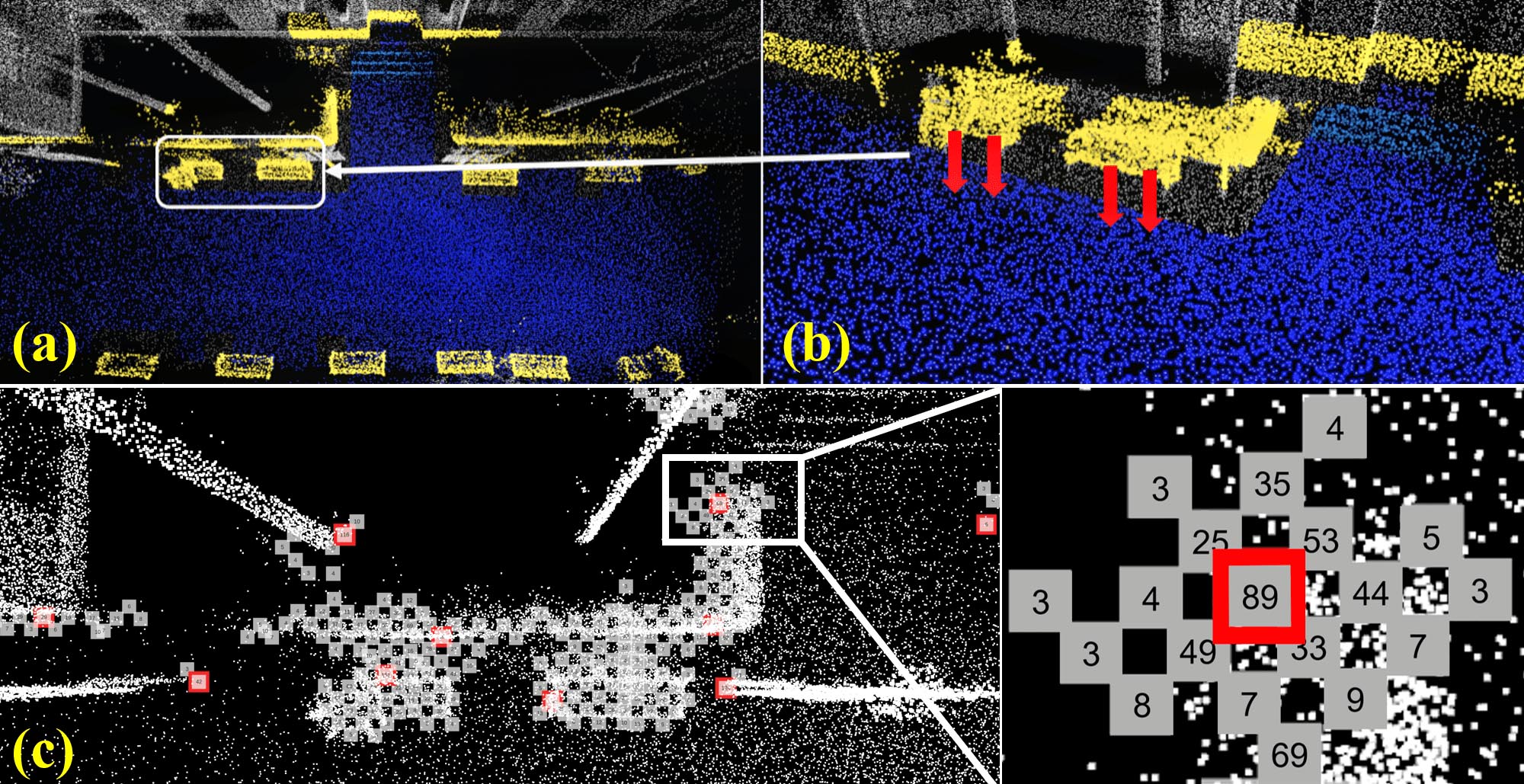}
    \caption{(a) Points in boundary voxels are colored in yellow. (b) The points are projected onto the adjacent planes (blue points). (c) The plane image where each pixel represents the maximum distance (in cm) of points in boundary voxels to the plane. If a point has the maximum pixel value in its 5*5 neighborhood, it  will be considered as the key points (red points). }
    \vspace{-0.3cm}
    \label{fig:plane_boundary}
\end{figure}

With the boundary voxels, we project their contained points to the respective plane (see Fig. \ref{fig:plane_boundary}(a) and Fig. \ref{fig:plane_boundary}(b)). For each plane, we create an image where the image plane coincides with the plane and each pixel represents the maximum distance of points contained in boundary voxels of the plane. Then, we select a point, which has the largest pixel value in its $5*5$ neighborhood, as a key point (see Fig. \ref{fig:plane_boundary}(c)). Each extracted key point corresponds to a 3D point in the input point cloud and could be attached with the normal of the plane it is extracted from.

With the extracted key points in a keyframe, we build a $k$-D tree and search 20 near neighbor points for each point to form the triangle descriptor. Redundant descriptors with the same side length will be eliminated. Each triangle descriptor contains three vertices, $\mathbf p_1$, $\mathbf p_2$ and $\mathbf p_3$, with projection normal vectors $\mathbf n_1$, $\mathbf n_2$ and $\mathbf n_3$. Besides, the vertices of the triangle are arranged according to the rule of side length in ascending order (see Fig. \ref{fig:triangle_descriptor}). We summarize that a triangle descriptor $\boldsymbol{\Delta}$ has the following contents:
\begin{itemize}
    \item $\mathbf p_1$, $\mathbf p_2$, $\mathbf p_3$: three vertices,
    \item $\mathbf n_1$, $\mathbf n_3$, $\mathbf n_3$: three projection normal vectors,
    \item $l_{12}$, $l_{23}$, $l_{13}$: three sides, and $l_{12} \leq l_{23} \leq l_{13}$,
    \item $\mathbf q$: the centriod of the triangle,
    \item $k$: the frame number corresponding to the descriptor.
\end{itemize}
In addition to descriptors, we will also save all $n$ planes ${\Pi}=(\boldsymbol{\pi}_1,\boldsymbol{\pi}_2,\ldots,\boldsymbol{\pi}_n)$ extracted from this keyframe for the following geometrical verification step.

% \begin{algorithm}
% \caption{Key point extraction}
% \SetKwInOut{Input}{\textbf{Input}}
% \SetKwInOut{Output}{\textbf{Output}}
% \Input{3D point cloud $\mathbf P_c$\;\\
%         Voxel size $ S$\;\\
%         Plane judgement threshold $\sigma$\\}
% \Output{Key points $\mathbf p_i \quad (i=1,2,...,n)$}
% Divide the point cloud $\mathbf P_c$ into voxels with voxel size $S$, get voxel list $\mathbf V_l$\\
% \For{each voxel $V_j$ in the Voxel list $\mathbf V_l$}{

% Fit a plane $\mathbf \pi_j$ with points in the voxel.\\
% Calculate the mean point-to-plane distance $d_j$.\\
% \If{$d_j<\sigma$}{
%  The points in the voxel are assumed to lie on the plane $\mathbf \pi_j$.\\
%  Save the plane $\mathbf \pi_j$ into plane list $\mathbf \Pi_l $ with plane center $\mathbf q_j$ and normal vector $\mathbf n_j$.\\
% }
% }
% \For{each plane $\pi_j$ in the plane list $\mathbf \Pi_l$}{
% Locate the voxel containing the plane $\pi_j$, search the nearby voxels.\\
% \uIf {Nearby voxels has the same plane}{
%     Merge the planes in the voxels.
% } \Else {
%   Add the voxel to a list of boundary voxels.
% }
% }
% \end{algorithm}

\subsection{Search of Loop Candidates}
Since hundreds of descriptors can be extracted from a keyframe, to quickly query and match descriptors, we use a Hash table to store all descriptors. We use six attributes with rotation and translation invariance in the descriptor to compute the hash key, which are side lengths $l_{12}$, $l_{23}$, $l_{13}$, and the dot product of the normal projection vector $\mathbf n_1 \cdot \mathbf n_2$, $\mathbf n_2 \cdot \mathbf n_3$, $\mathbf n_1 \cdot \mathbf n_3$ respectively. Descriptors with all six similar attributes will have the same hash key and hence be stored in the same container. For a query keyframe, we extract all its descriptors as detailed in Sec. \ref{method:descriptor}. For each descriptor $\boldsymbol{\Delta}_i$, we calculate its hash key, locate it to the corresponding container in the Hash table and vote once for the keyframes that have a descriptor in this container. The matching process finishes when all descriptor $\boldsymbol{\Delta}_i$ in the query keyframe are processed. keyframes with the top 10 votes will be selected as candidates with matched descriptors saved for the use of the loop detection step.

{\it Remark 1:} Since boundary points are projected to the plane extracted from the 3D point cloud instead of from the range image, such as in \cite{pointFeature2010icra}, the extracted key point is invariant to the view angle change. Moreover, the six descriptor attributes are also invariant to any rigid transformation. Hence, the overall method is rotation and translation invariance. 

{\it Remark 2:} Thanks to the ordering of the triangle side length and the stability of the triangle, two triangles are assured to be the same if and only if the length of their ordered sides are equal, without enumerating the side correspondence. 

% \begin{equation}
%     NPD = \frac{1}{3} \sum_{k=1}^3 \frac{|PD_{ki}- PD_{kj}|}{max(PD_{ki},PD_{kj})}
% \end{equation}

% The matching process finishes when all the candidate containers are explored and returns a list of matched descriptor pair with NPD.  Then we sort the keyframe numbers according to the total NPD size from large to small, and select the first 10/50 keyframes as candidate frames. 

\subsection{Loop Detection}
\label{sec:loop_detection}
When given a loop candidate keyframe, we perform geometrical verification to eliminate the false detection due to incorrect descriptor matching pairs.
Since the shape of the triangle is uniquely determined after the side length is determined, once $\mathbf \Delta_a$ is matched to $\mathbf \Delta_b$, their vertices $(\mathbf p_{a_1}, \mathbf p_{a_{2}}, \mathbf p_{a_3})$ and $(\mathbf p_{b_1}, \mathbf p_{b_2}, \mathbf p_{b_3})$ naturally match. Then with this point correspondence, we can easily calculate the relative transformation $\mathbf T=(\mathbf R,\mathbf t)$ between these two keyframes through Singular Value Decomposition (SVD):
\begin{equation}
\abovedisplayshortskip=3pt
\belowdisplayshortskip=3pt
\abovedisplayskip=3pt
\belowdisplayskip=3pt
\begin{aligned}
      &  \mathbf H = \sum\nolimits_{i=1}^3 (\mathbf p_{a_i}-\mathbf q_{a})(\mathbf p_{b_i}-\mathbf{q}{_b}) \\
      &  [\mathbf U,\ \mathbf S,\ \mathbf V] = \mathtt{SVD}(\mathbf H)\\
      & \mathbf R = \mathbf{VU}^T , \ \mathbf t = -\mathbf{R} * \mathbf q_a + \mathbf q_b.
\end{aligned}
\end{equation}
To increase the robustness, we use RANSAC\cite{fischler1981ransac} to find the transformation that maximizes the number of correctly matched descriptors.
% The transformation can be used for further geometric verification with the plane groups ${\Pi}=(\boldsymbol{\pi}_1,\boldsymbol{\pi}_2,..\boldsymbol{\pi}_n)$  extracted in \ref{method:descriptor} to improve the accuracy of loop detection. 

Based on this transformation, we calculate the plane overlap between the current frame and the candidate frame for geometrical verification. Let a center point $\mathbf g$ and a normal vector $\mathbf u$ represent a plane $\boldsymbol{\pi}$ in a voxel. Denote the plane group of the current frame be ${}^B \Pi=[({}^B \mathbf g_{1}, {}^B \mathbf u_{1}),...({}^B \mathbf g_{n},{}^B \mathbf u_{n})])$, the plane group of the candidate frame be ${}^C \Pi=[({}^C \mathbf g_{1}, {}^C \mathbf u_{1}),...({}^C \mathbf g_{m},{}^C \mathbf u_{m})])$, and the rigid-body transformation be ${}^C_B \mathbf T=({}^C_B \mathbf R, {}^C_B \mathbf t) \in SE(3))$, where $n$ is the number of planes in the current frame and $m$ is the number of planes in the candidate frame. We construct a $k$-D tree ($k=3$) with the center points $({}^C \mathbf g_1,{}^C \mathbf g_2,..., {}^C \mathbf g_m )$ from the ${}^C \Pi$. Then for each plane center point ${}^B \mathbf g_i \ (i=1,2,...,n) \in {}^B \Pi$. We first transform ${}^B \mathbf g_i$ by the transformation ${}^C_B \mathbf T$, then search a nearest point ${}^C \mathbf g_j$ in the $k$-D tree, and judge whether the two planes coincide by the difference in normal vector and the point-to-plane distance:
\begin{equation}
\label{eq:optimize}
\abovedisplayshortskip=3pt
\belowdisplayshortskip=3pt
\abovedisplayskip=3pt
\belowdisplayskip=3pt
\begin{aligned}
        \|{}^C_B\mathbf R {}^B\mathbf u_i-{}^C \mathbf u_j \|_2 <\sigma_n \\ 
        {}^C \mathbf u_j^T({}^C_B\mathbf T {}^B \mathbf g_i-{}^C\mathbf g_j) < \sigma_d,
\end{aligned}
\end{equation}
where $\sigma_n$ and $\sigma_d$ are preset hyperparameters to determine whether planes overlap or not. If a pair of planes satisfies the normal vector and point-to-distance constraints in equation (\ref{eq:optimize}), the pair of planes are coinciding. After checking all planes of the current frame, we calculate the percent of plane coincidence ($N_c$):
	\begin{equation}
 \abovedisplayshortskip=3pt
\belowdisplayshortskip=3pt
\abovedisplayskip=3pt
\belowdisplayskip=3pt
		N_c = \frac{N_{\mathtt{coincide}}}{N_{\mathtt{sum}}} \times 100 \%,
	\end{equation}
where $N_{\mathtt{coincide}}$ is the number of coninciding planes and $N_{\mathtt{sum}}$ is the number of all planes of the current frame. If the $N_c$ of the current frame and the candidate frame exceeds a certain threshold $\sigma_{\mathtt{pc}}$, we finally consider it to be a valid loop detection. It is worth noting that geometric verification based on planes is much more efficient than the ICP-based methods since the number of planes is much less than the number of point clouds. Besides, we can further optimize the normal vector difference and point-to-plane distance in equation (\ref{eq:optimize}) to obtain a more accurate transformation for loop correction, which can be easily implemented using Ceres-Solver\cite{ceres-solver}. We defined this optimization process as STD-ICP and the performance of STD-ICP will be verified in experiments.

\section{Experiments}
In this section, to verify the effectiveness, robustness and adaptability of our method, we evaluate our algorithm in different scenarios (urban, indoor and unstructured environments) with different types of LiDAR (mechanical spinning LiDARs and solid state LiDARs). In each experiment, we compare our method with state-of-the-art counterparts. All experiments are carried out on the same system with an Intel i7-11700k @ 3.6 $\rm GHz$ with 16 $\rm{GB}$ memory.
\subsection{Benchmark Evaluation \label{exp:benchmark}}
In this experiment, we evaluate our method on the open urban dataset including KITTI odometry dataset \cite{KITTI}, NCLT dataset \cite{NCLT} and Complex Urban dataset \cite{complex_urban_dataset}. All data were collected in urban environments using mechanical spinning LiDARs with different scanning lines. We compare our method with two other global descriptors: Scan Context \cite{kim2018scan} and M2DP \cite{m2dp}. We accumulate every 10 frames into a keyframe for these datasets. If the ground truth pose distance between the query keyframe and the matched keyframe is less than 20m, the detection is considered a true positive.

For the implementation, we run our algorithm on all datasets with the same parameters, where the voxel size is 1m, the plane judgement threshold $\sigma_1$ and $\sigma_2$ are 0.01 and 0.05, respectively, the normal different threshold $\sigma_n$ is 0.2 and the point-to-plane distance threshold $\sigma_d$ is 0.3m. For Scan Context \cite{kim2018scan} and M2DP \cite{m2dp}, we directly use the results presented in the original paper \cite{kim2018scan}.

% \begin{table}[]
%     \label{tab:urban_dataset}
%     \centering
%     \begin{tabular}{ccccc}
%     \toprule
%     Dataset & Sequence index & \# of true loops & \# of all nodes & Length (m)\\
%     \midrule
%     \multirow{4}{*}{KITTI} & 00 & 94 & 450 & 3714\\
%     & 02 & 38 & 466 & 4268 \\
%     & 05 & 50 & 267 & 2223 \\
%     & 08 & 49 & 399 & 3225 \\ 
%     \hline
%     \multirow{4}{*}{NCLT} & 20120526 & 312 & 1175 & 6345\\
%     & 20120820 & 212 & 1121 & 6018\\
%     & 20120928 & 251 & 1045 & 5579\\
%     & 20130405 & 130 & 858 & 4530\\
%     \bottomrule
%     \end{tabular}
%     \caption{LOOP NODES AND ALL NODES OF CHOOSEN DATASET}
%     \label{tab:my_label}
% \end{table}

\begin{figure*}[ht]
    \centering
    \vspace{-0.4cm}
    \setlength{\abovecaptionskip}{-0.1cm}
    \includegraphics[width=0.95\linewidth]{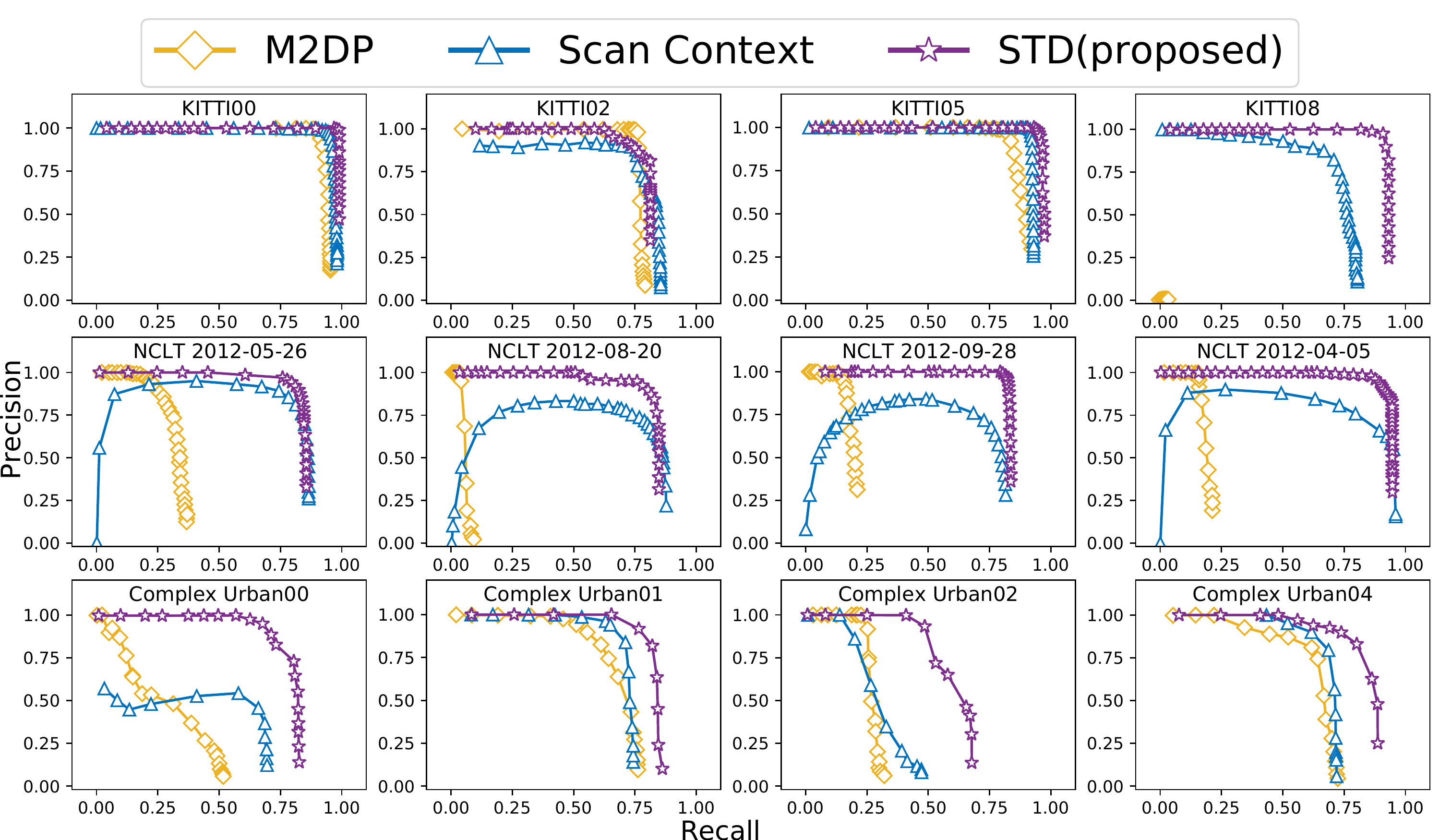}
    \caption{Precision-Recall curves on KITTI, NCLT and Complex Urban datasets .}
    \vspace{-0.4cm}
    \label{fig:urban_pr_curve}
\end{figure*}

\subsubsection{Precision-Recall Evaluation\label{exp:pr_curve}}
\
\par
\begin{figure}[h]
    \centering
    \setlength{\abovecaptionskip}{-0.1cm}
    \includegraphics[width=1\linewidth]{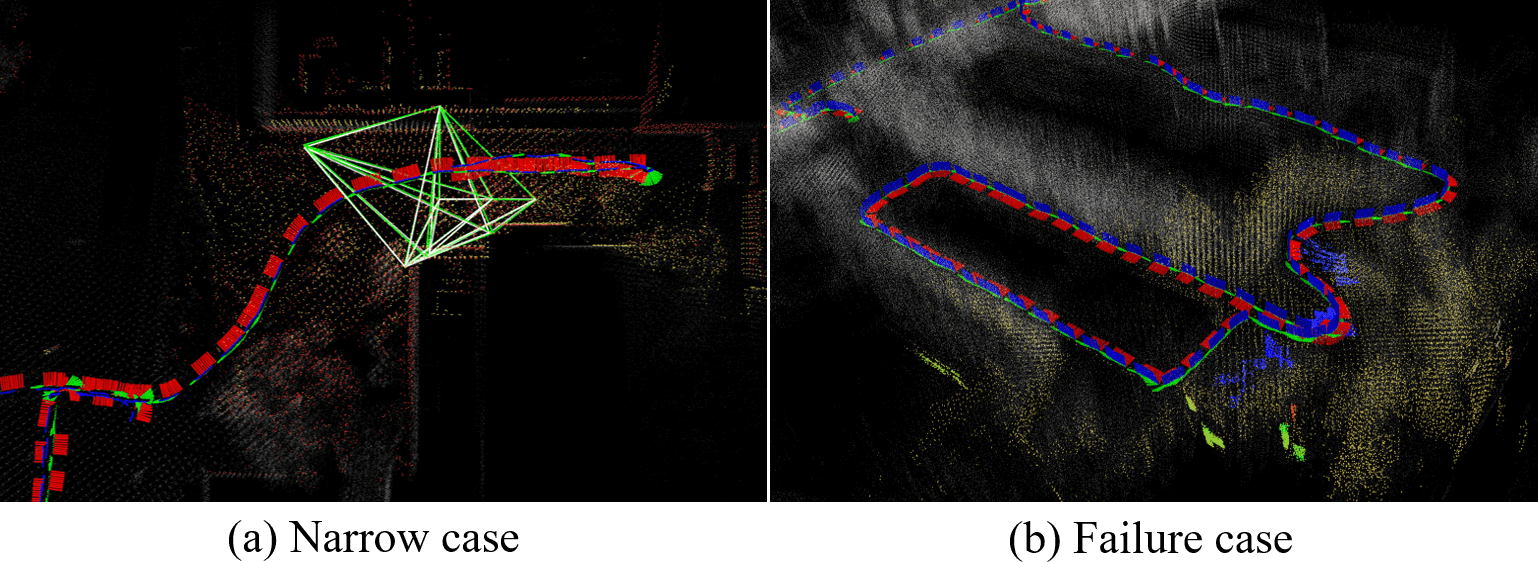}
    \caption{A challenge case in narrow scenario and a failure case of STD.}
    \vspace{-0.6cm}
    \label{fig:nclt_case}
\end{figure}

We evaluate the performance of STD by the precision-recall curve when vary plane coincide threshold $\sigma_{\mathtt{pc}}$ as shown in Fig. \ref{fig:urban_pr_curve}. Since Scan Context-50 performs better than Scan Context-10 in most scenarios, we only show the result from Scan Context-50. From the results, STD outperforms Scan Context and M2DP in almost all datasets. As stated in \cite{kim2018scan}, their method does not perform as well in narrow scenarios where variation in vertical height is less significant. However, our method is not limited to the height of the scene, and a successful loop detection in such a scene is shown in Fig. \ref{fig:nclt_case} (a). Our method performs poorly only when the structure or planes of the scene are particularly sparse because the key points extracted in such scenes will be scarce. A typical failure example is shown in Fig. \ref{fig:nclt_case} (b). Both cases are from the NCLT dataset.

\subsubsection{Run Time Evaluation}
\
\par
\begin{figure*}[ht]
		% \centering
		\begin{minipage}{0.70\linewidth}
			\vspace{-0.4cm}
			\includegraphics[width=1\linewidth]{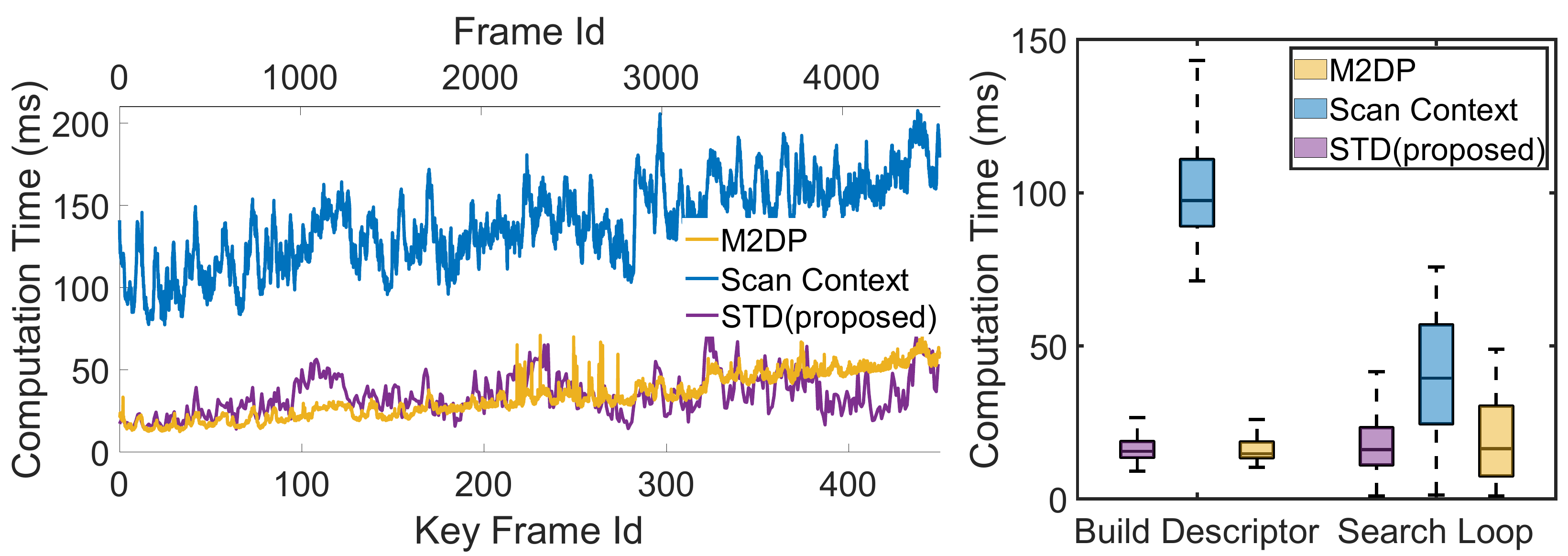}
			\vspace{-0.8cm}
			\caption{Time Evaluation on KITTI00}
			\label{fig:time_evaluation}
		\end{minipage}
		\hspace{0.01cm}
		\begin{minipage}{0.28\linewidth}
			\includegraphics[width=1\linewidth]{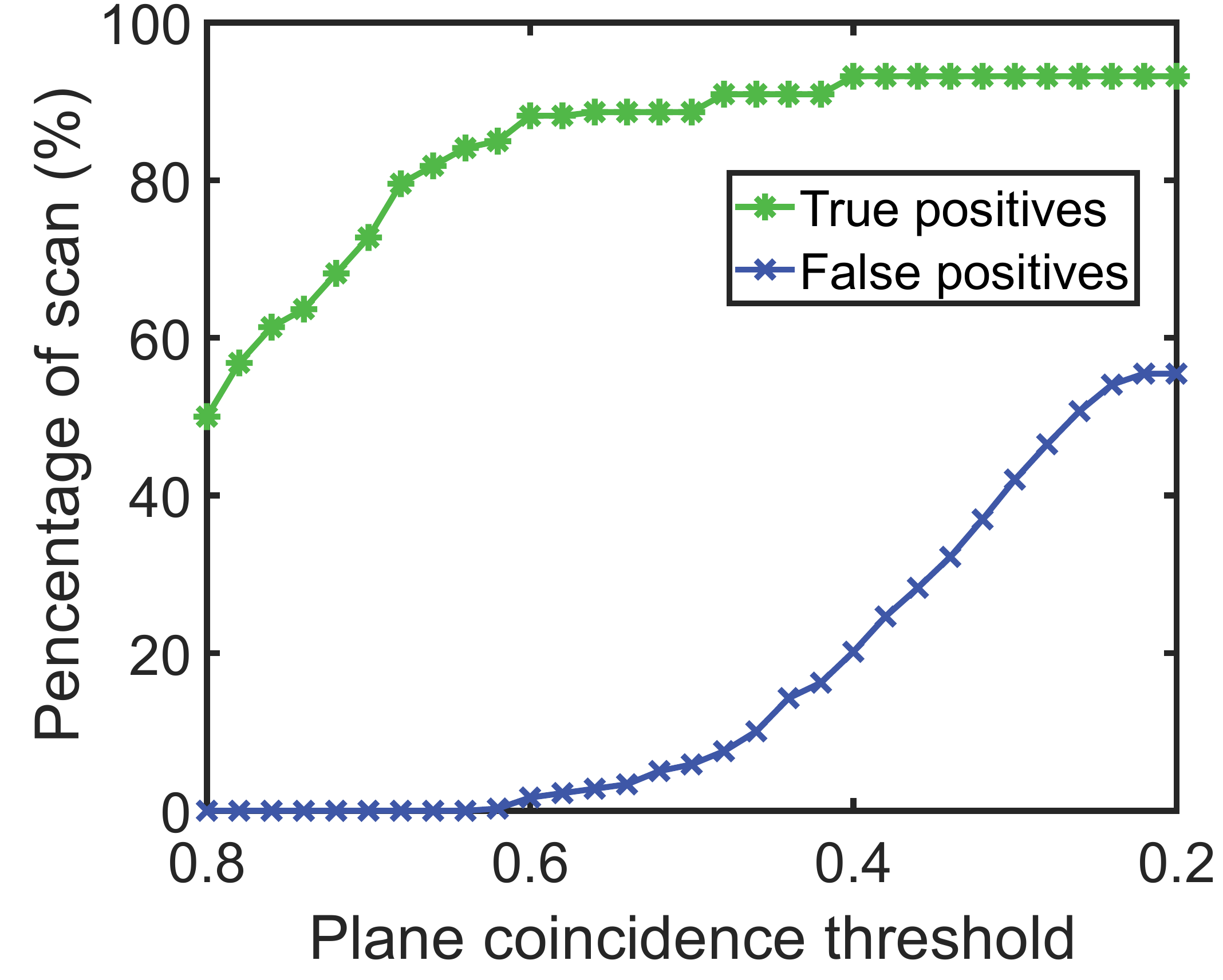}
			\vspace{-0.8cm}
    			\caption{The effect of the plane coincidence threshold $\sigma_{\mathtt{pc}}$ on the true positive rate and false positive rate on KITTI08.}
			\label{fig:pc_threshold}
		\end{minipage}
	\end{figure*}

We record the computation time on KITTI00 for all methods, as shown in Fig. \ref{fig:time_evaluation}. For M2DP \cite{m2dp}, we test with their open-sourced MATLAB code with the default parameter. For Scan Context \cite{kim2018scan}, we modified their MATLAB code (add 8 Scan Context augmentations) to obtain the results in Sec. \ref{exp:pr_curve}. Point cloud downsampling with a $0.5 \ m^3$ voxel grid is used for all methods. As shown in Fig. \ref{fig:time_evaluation}, the time consumption per frame in Scan Context and M2DP linearly increases with the number of frames in the library, while our method does not have such a linear growing trend. That is mainly due to our use of a Hash table as the database to store descriptors, which avoids building a $k$-D tree for historical descriptors like M2DP and Scan Context do. Overall, the computation time of STD is similar with M2DP, while it processing 10 times more points than M2DP. Scan Context uses augmented descriptors, which increases time consumption in both descriptor build and search loop.
% We divide the whole place identification task into 3 step: descriptor extraction, candidate frame search and loop detection. Fig. \ref{fig:time_eval} records the processing time of each step of loop frames and non-loop frames on KITTI00 dataset. It can be seen that the extraction of descriptors is very fast, the average time is only 10ms, and the maximum time is not more than 20ms. Compared with non-loop frame, loop frame has more matched descriptor pairs, so the rigid-body transform solution and geometric verification based on RANSAC are more time-consuming than non-loop frame, but the total time is still within an acceptable range.

\subsubsection{Plane Coincidence Threshold Selection}
\
\par
As can be seen from Fig. \ref{fig:urban_pr_curve}, Precision-Recall curves of STD always decline from precision equal to 1, mainly due to the selection of the plane coincide threshold $\sigma_{\mathtt{pc}}$. When a relatively large  $\sigma_{\mathtt{pc}}$ is given, only the loop with large point cloud overlap will be selected, which is $100\%$ accurate in the urban dataset we use. When the threshold decreases, more loops with smaller overlap will be selected, introducing possible false positives. We record the true and false positive rates corresponding to different $\sigma_{\mathtt{pc}}$ of STD on Kitti08 in Fig. \ref{fig:pc_threshold}. It can be seen from the figure that $0.5\sim 0.6$ is a good trade-off value.
% \begin{figure}[h]
%     \centering

% \end{figure}

\subsubsection{Localization Evaluation}
\
\par
\begin{figure*}[ht]
		%\centering
		\begin{minipage}{0.70\linewidth}
			\vspace{-0.8cm}
			\includegraphics[width=1\linewidth]{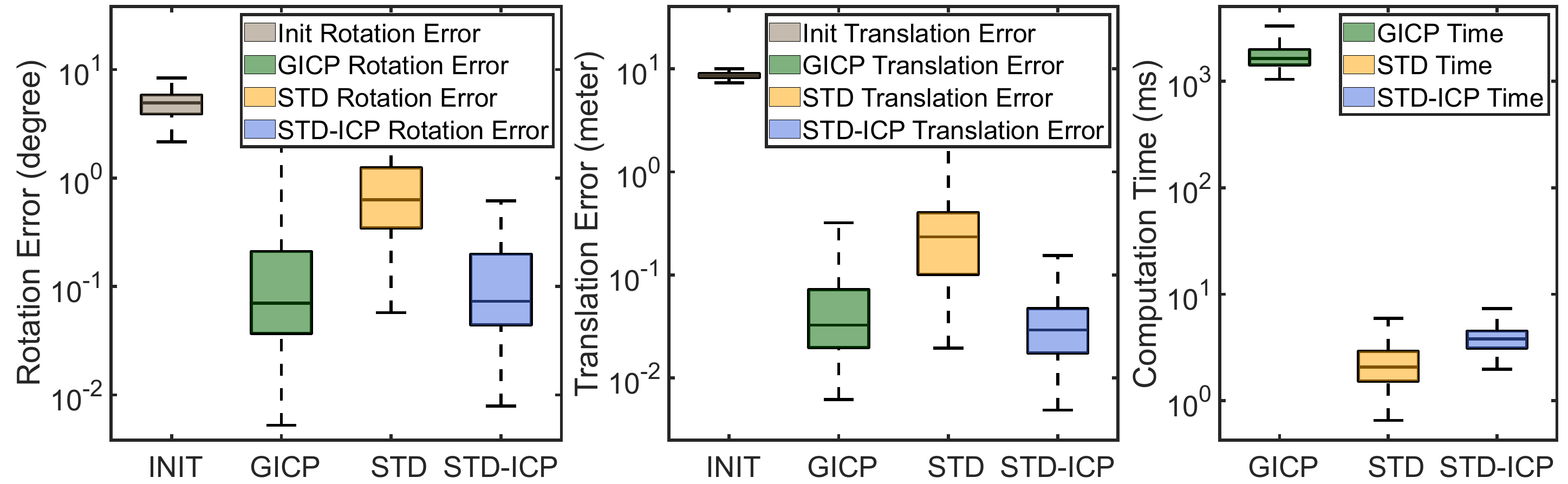}
			\vspace{-0.8cm}
			\caption{Pose error and computation time of GICP, STD and STD-ICP.}
			\label{fig:localization}    
		\end{minipage}
		\hspace{0.1cm}
		\begin{minipage}{0.28\linewidth}
			\vspace{-0.2cm}
			\includegraphics[width=1\linewidth]{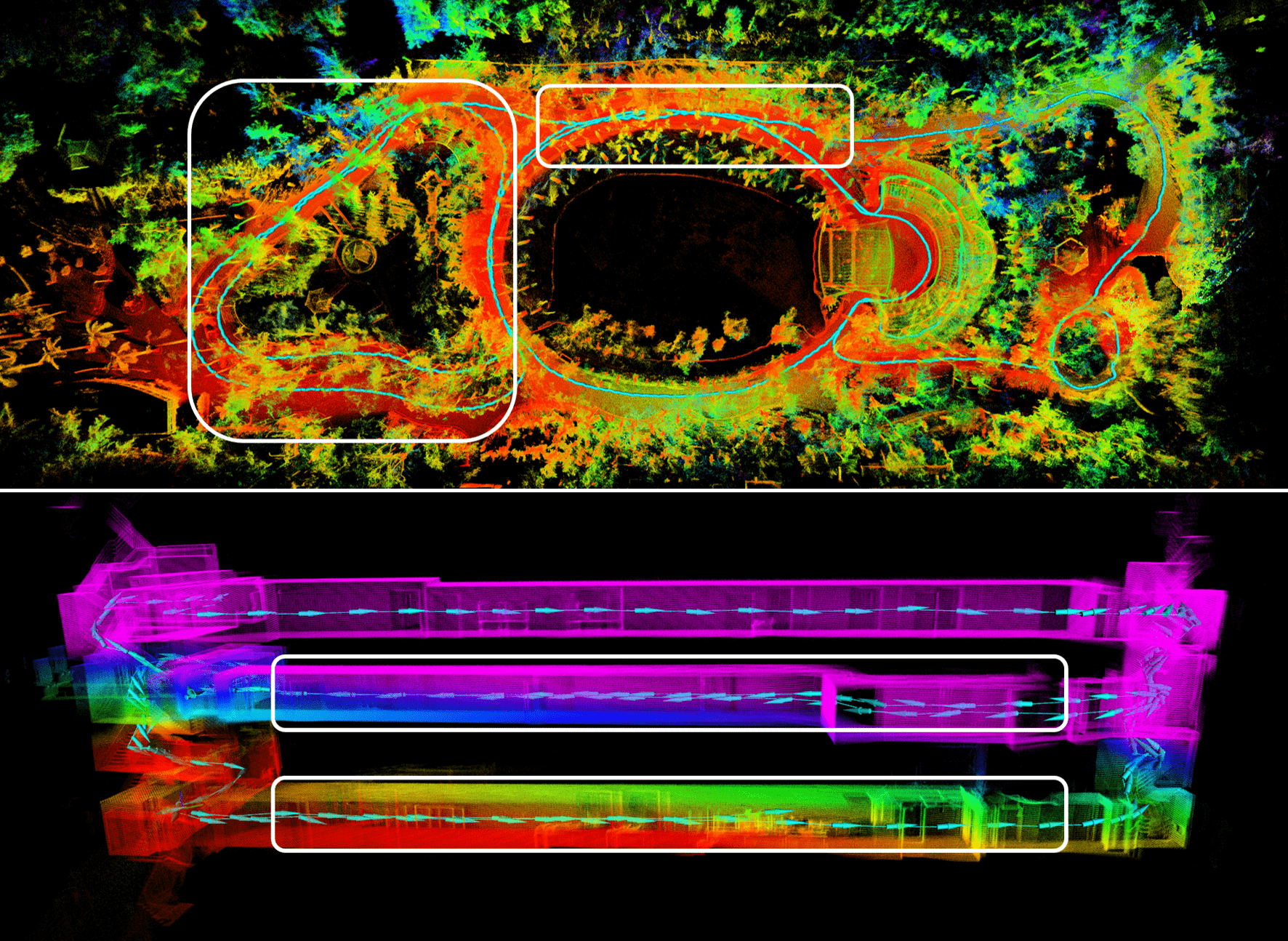}
			\vspace{-0.8cm}
			\caption{The fine point cloud map in park and indoor environments. Loop nodes are indicated by white boxes.}
			\label{fig:avia_scene}
		\end{minipage}		
		\vspace{-0.6cm}
	\end{figure*}

% \begin{figure}[ht]
%     \centering
%     \includegraphics[width=1\linewidth]{figures/rmse_time.png}
%     \caption{RMSE and computation time in localization evaluation.}
%     \label{fig:transformation}
% \end{figure}
Some other descriptors \cite{kim2018scan,overlap2020rss} can estimate the yaw angle between the loop frame and the candidate frame while performing loop detection. Our proposed descriptor has further improved this function since we can provide the relative transformation of all six degrees of freedom between the loop frame and the candidate frame without extra calculation. To verify this, we conduct the experiment on the loop nodes of KITTI00. For each loop node, we set the transform relative to matched frame a random initial value uniformly drawn from a neighborhood ($\pm 5^{\circ}$ in each axis of rotation and $\pm 5$m in each axis of translation) of the ground truth value. Fig. \ref{fig:localization} shows the error and computation time of GICP, STD and STD-ICP. STD-ICP can achieve similar accuracy as GICP with less variance in both rotation and translation. This is because STD provides a good initial value for STD-ICP, while GICP is likely to have a local optimum in loop nodes with less overlap. In addition, STD and STD-ICP take much less time than GICP, only less than $1\%$ of GICP. This is because the number of planes (hundreds) is very small compared to the size of the point cloud (more than 100K).

%test

\subsection{Applicability to Other Types of LiDARs}
In this experiment, to evaluate the adaptability and applicability of STD in different environments and using different LiDARs, we conduct experiments with Livox series solid-state LiDARs in urban, unstructured and indoor environments. For the urban environment, we choose the 	KA\_Urban\_East dataset, which is collected by Livox Horizon LiDAR and open-sourced in LiLi-OM \cite{liliom}. For the unstructured environment experiment, we collect two groups of loop data inside a park filled with trees. For the indoor environment, we collect the loop data in a multi-floor building. These two datasets are collected by Livox Avia LiDAR. Since Scan Context \cite{kim2018scan} is not compatible with Livox solid-state LiDARs, so we only compare STD with M2DP \cite{m2dp}. 
For the implementation, we use the default parameters of the available codes for M2DP \cite{m2dp}. For STD, the voxel size used in indoor is adjust to 0.5m while other parameters remain the same as in Sec. \ref{exp:benchmark}.

For the ground truth calculation, we first use a LiDAR-Inertial Odometry to get a rough map, then use loop detection and pose graph \cite{kaess2008isam, kaess2012isam2} to get a fine map, which is used as the true value to select the ground truth loop nodes. Because the number of the point cloud in a single frame of Livox LiDAR is sparser than that of spinning LiDAR, we accumulate every 20 frames into a keyframe. Based on this, if the ground truth pose distance between the query keyframe and the matched keyframe is less than 20m for outdoor and 4m for indoor, the detection is considered as a true positive. We show the fine point cloud map and loop nodes for the park and indoor environments in Fig. \ref{fig:avia_scene}. 
\subsubsection{Result Analysis}
\
\par
\begin{figure}[ht]
\vspace{-0.1cm}
    \centering
    \setlength{\abovecaptionskip}{-0.1cm}
    \includegraphics[width=1\linewidth]{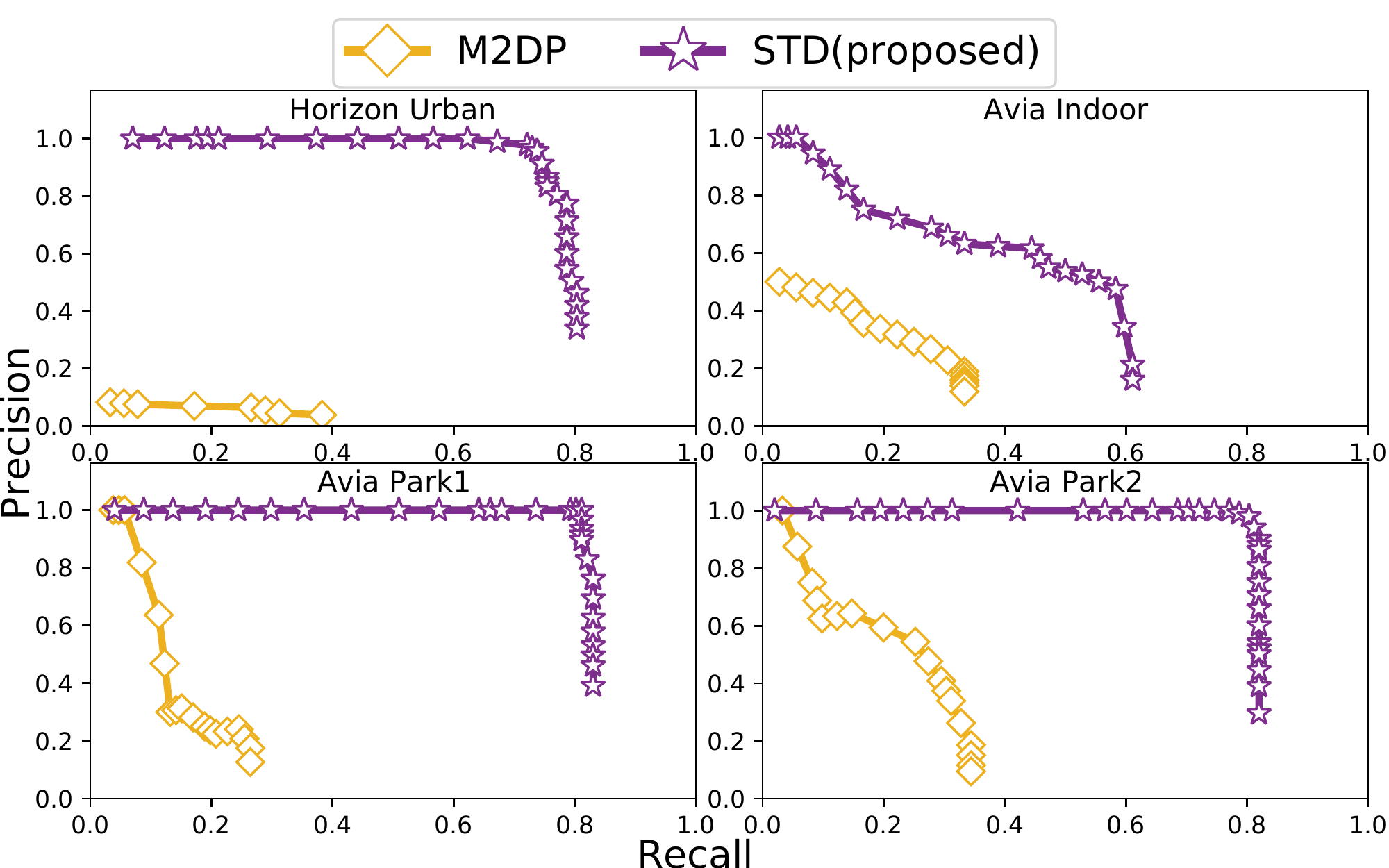}
    \caption{Precision-Recall curves on Livox LiDAR dataset.}
     \vspace{-0.4cm}
    \label{fig:avia_pr_curve}
\end{figure}

The Precision-Recall curves for STD and M2DP \cite{m2dp} are  plotted in Fig. \ref{fig:avia_pr_curve}. From the figure, we can see that M2DP performs poorly on the Livox dataset, while STD achieves similar performance as \ref{exp:benchmark} on the Livox dataset, except indoor dataset. This is mainly because the corridors of each floor of the building are very similar, resulting in relatively low precision and recall. However, we can still provide a certain number of valid loop nodes for loop correction so that LiDAR loop closure can be applied to indoor mappings, such as multi-floor parking lots, museums, etc.

% Since the construction of STD does not make any assumptions about the environment and LiDAR installation,our method could also be applied to Livox series solid state LiDARs which have fixed orientation FoV and non-repetitive scanning without any modification. 

% \begin{figure}[h]
%     \centering
%     \includegraphics[width=1\linewidth]{avia_scene.png}
%     \caption{The fine point cloud map in park and indoor environments. Loop nodes are indicated by white boxes.}
%     \label{fig:avia_scene}
% \end{figure}

\section{Conclusion}
This paper proposes a triangle-based global descriptor  {\it STD}. An efficient key point extraction algorithm based on plane detection and boundary projection is proposed to extract key points with geometrical features. These key points form the triangle descriptors with their neighbors. This combination greatly improves the rotation and translation invariance of descriptors. Besides, the stability and uniqueness of triangles make this descriptor naturally suitable for similarity comparison in place recognition. To speed up the querying and matching of the descriptor, we employ a Hash table as the database to store all historical descriptors, which avoids building a $k$-D tree in loop searching. Compared with other global descriptors, STD not only performs better on public datasets but also shows greater adaptability to different environments and LiDAR types.

%In future work, we plan to exploit more local information on the key point since we only use the position and the projection normal information of the key points. In addition, we find that there are some descriptors redundant in the database, which will affect the recall accuracy and efficiency. How to remove the redundancy to further improve the uniqueness of the descriptor is also a direction.
% \appendices
% \section{}

% \section{}


% Acknowledgment 可以等review之后，提交camera ready的版本时加
\section*{Acknowledgment}
This project is supported by DJI under the grant 17206421 and in part by Shenzhen Science and Technology Project (\text{JSGG20211029095803004,\ JSGG20201103100401004}).
The authors would like to thank DJI Co., Ltd\footnote{\url{https://www.dji.com}} for providing devices and research found.

\bibliographystyle{IEEEtran}
\bibliography{root}

\begin{thebibliography}{10}
\providecommand{\url}[1]{#1}
\csname url@samestyle\endcsname
\providecommand{\newblock}{\relax}
\providecommand{\bibinfo}[2]{#2}
\providecommand{\BIBentrySTDinterwordspacing}{\spaceskip=0pt\relax}
\providecommand{\BIBentryALTinterwordstretchfactor}{4}
\providecommand{\BIBentryALTinterwordspacing}{\spaceskip=\fontdimen2\font plus
\BIBentryALTinterwordstretchfactor\fontdimen3\font minus
  \fontdimen4\font\relax}
\providecommand{\BIBforeignlanguage}[2]{{%
\expandafter\ifx\csname l@#1\endcsname\relax
\typeout{** WARNING: IEEEtran.bst: No hyphenation pattern has been}%
\typeout{** loaded for the language `#1'. Using the pattern for}%
\typeout{** the default language instead.}%
\else
\language=\csname l@#1\endcsname
\fi
#2}}
\providecommand{\BIBdecl}{\relax}
\BIBdecl

\bibitem{lin2020loam}
J.~Lin and F.~Zhang, ``Loam$\_$livox: A fast, robust, high-precision lidar
  odometry and mapping package for lidars of small fov,'' in \emph{2020 IEEE
  International Conference on Robotics and Automation (ICRA)}.\hskip 1em plus
  0.5em minus 0.4em\relax IEEE, 2020, pp. 3126--3131.

\bibitem{appearance-loop2013tro}
M.~Labbé and F.~Michaud, ``Appearance-based loop closure detection for online
  large-scale and long-term operation,'' \emph{IEEE Transactions on Robotics},
  vol.~29, no.~3, pp. 734--745, 2013.

\bibitem{lin2019fast}
J.~Lin and F.~Zhang, ``A fast, complete, point cloud based loop closure for
  lidar odometry and mapping,'' \emph{arXiv preprint arXiv:1909.11811}, 2019.

\bibitem{1-year2019ral}
G.~Kim, B.~Park, and A.~Kim, ``1-day learning, 1-year localization: Long-term
  lidar localization using scan context image,'' \emph{IEEE Robotics and
  Automation Letters}, vol.~4, no.~2, pp. 1948--1955, 2019.

\bibitem{rgb-relocalization}
B.~Glocker, J.~Shotton, A.~Criminisi, and S.~Izadi, ``Real-time rgb-d camera
  relocalization via randomized ferns for keyframe encoding,'' \emph{IEEE
  Transactions on Visualization and Computer Graphics}, vol.~21, no.~5, pp.
  571--583, 2015.

\bibitem{decentralize2020iros}
J.~Lin, X.~Liu, and F.~Zhang, ``A decentralized framework for simultaneous
  calibration, localization and mapping with multiple lidars,'' in \emph{2020
  IEEE/RSJ International Conference on Intelligent Robots and Systems (IROS)},
  2020, pp. 4870--4877.

\bibitem{orbslam}
R.~Mur-Artal, J.~M.~M. Montiel, and J.~D. Tardós, ``Orb-slam: A versatile and
  accurate monocular slam system,'' \emph{IEEE Transactions on Robotics},
  vol.~31, no.~5, pp. 1147--1163, 2015.

\bibitem{qin2018vins}
T.~Qin, P.~Li, and S.~Shen, ``Vins-mono: A robust and versatile monocular
  visual-inertial state estimator,'' \emph{IEEE Transactions on Robotics},
  vol.~34, no.~4, pp. 1004--1020, 2018.

\bibitem{r2live}
J.~Lin, C.~Zheng, W.~Xu, and F.~Zhang, ``R$^2$live: A robust, real-time,
  lidar-inertial-visual tightly-coupled state estimator and mapping,''
  \emph{IEEE Robotics and Automation Letters}, vol.~6, no.~4, pp. 7469--7476,
  2021.

\bibitem{r3live}
J.~Lin and F.~Zhang, ``R$^3$live: A robust, real-time, rgb-colored,
  lidar-inertial-visual tightly-coupled state estimation and mapping package,''
  in \emph{2022 International Conference on Robotics and Automation (ICRA)},
  2022, pp. 10\,672--10\,678.

\bibitem{zhu2020camvox}
Y.~Zhu, C.~Zheng, C.~Yuan, X.~Huang, and X.~Hong, ``Camvox: A low-cost and
  accurate lidar-assisted visual slam system,'' 2020.

\bibitem{yang2022lidarVelocity}
W.~Yang, Z.~Gong, B.~Huang, and X.~Hong, ``Lidar with velocity: Correcting
  moving objects point cloud distortion from oscillating scanning lidars by
  fusion with camera,'' \emph{IEEE Robotics and Automation Letters}, vol.~7,
  no.~3, pp. 8241--8248, 2022.

\bibitem{yuanpixel}
C.~Yuan, X.~Liu, X.~Hong, and F.~Zhang, ``Pixel-level extrinsic self
  calibration of high resolution lidar and camera in targetless environments,''
  \emph{IEEE Robotics and Automation Letters}, vol.~6, no.~4, pp. 7517--7524,
  2021.

\bibitem{fastlio2}
W.~Xu, Y.~Cai, D.~He, J.~Lin, and F.~Zhang, ``Fast-lio2: Fast direct
  lidar-inertial odometry,'' \emph{IEEE Transactions on Robotics}, 2022.

\bibitem{r3live_pp}
J.~Lin and F.~Zhang, ``R$^3$live++: A robust, real-time, radiance
  reconstruction package with a tightly-coupled lidar-inertial-visual state
  estimator,'' \emph{arXiv preprint arXiv:2209.03666}, 2022.

\bibitem{pointFeature2010icra}
B.~Steder, G.~Grisetti, and W.~Burgard, ``Robust place recognition for 3d range
  data based on point features,'' in \emph{2010 IEEE International Conference
  on Robotics and Automation}, 2010, pp. 1400--1405.

\bibitem{keypointBosse2013icra}
M.~Bosse and R.~Zlot, ``Place recognition using keypoint voting in large 3d
  lidar datasets,'' in \emph{2013 IEEE International Conference on Robotics and
  Automation}, 2013, pp. 2677--2684.

\bibitem{PFH}
R.~B. Rusu, N.~Blodow, Z.~C. Marton, and M.~Beetz, ``Aligning point cloud views
  using persistent feature histograms,'' in \emph{2008 IEEE/RSJ International
  Conference on Intelligent Robots and Systems}, 2008, pp. 3384--3391.

\bibitem{SURF}
H.~Bay, T.~Tuytelaars, and L.~Van~Gool, ``Surf: Speeded up robust features,''
  in \emph{European conference on computer vision}.\hskip 1em plus 0.5em minus
  0.4em\relax Springer, 2006, pp. 404--417.

\bibitem{SHOT}
S.~Salti, F.~Tombari, and L.~Di~Stefano, ``Shot: Unique signatures of
  histograms for surface and texture description,'' \emph{Computer Vision and
  Image Understanding}, vol. 125, pp. 251--264, 2014.

\bibitem{ndt}
M.~Magnusson, H.~Andreasson, A.~N{\"u}chter, and A.~J. Lilienthal, ``Automatic
  appearance-based loop detection from three-dimensional laser data using the
  normal distributions transform,'' \emph{Journal of Field Robotics}, vol.~26,
  no. 11-12, pp. 892--914, 2009.

\bibitem{m2dp}
L.~He, X.~Wang, and H.~Zhang, ``M2dp: A novel 3d point cloud descriptor and its
  application in loop closure detection,'' in \emph{2016 IEEE/RSJ International
  Conference on Intelligent Robots and Systems (IROS)}.\hskip 1em plus 0.5em
  minus 0.4em\relax IEEE, 2016, pp. 231--237.

\bibitem{kim2018scan}
G.~Kim and A.~Kim, ``Scan context: Egocentric spatial descriptor for place
  recognition within 3d point cloud map,'' in \emph{2018 IEEE/RSJ International
  Conference on Intelligent Robots and Systems (IROS)}.\hskip 1em plus 0.5em
  minus 0.4em\relax IEEE, 2018, pp. 4802--4809.

\bibitem{forest}
G.~V. Nardari, A.~Cohen, S.~W. Chen, X.~Liu, V.~Arcot, R.~A.~F. Romero, and
  V.~Kumar, ``Place recognition in forests with urquhart tessellations,''
  \emph{IEEE Robotics and Automation Letters}, vol.~6, no.~2, pp. 279--286,
  2021.

\bibitem{jiang2d}
B.~Jiang, Y.~Zhu, and M.~Liu, ``A triangle feature based map-to-map matching
  and loop closure for 2d graph slam,'' in \emph{2019 IEEE International
  Conference on Robotics and Biomimetics (ROBIO)}, 2019, pp. 2719--2725.

\bibitem{segmatch2017icra}
R.~Dub{\'e}, D.~Dugas, E.~Stumm, J.~Nieto, R.~Siegwart, and C.~Cadena,
  ``Segmatch: Segment based place recognition in 3d point clouds,'' in
  \emph{2017 IEEE International Conference on Robotics and Automation
  (ICRA)}.\hskip 1em plus 0.5em minus 0.4em\relax IEEE, 2017, pp. 5266--5272.

\bibitem{overlap2020rss}
X.~Chen, T.~L\"abe, A.~Milioto, T.~R\"ohling, O.~Vysotska, A.~Haag, J.~Behley,
  and C.~Stachniss, ``{OverlapNet: Loop Closing for LiDAR-based SLAM},'' in
  \emph{Proceedings of Robotics: Science and Systems (RSS)}, 2020.

\bibitem{yuan2022voxelMap}
C.~Yuan, W.~Xu, X.~Liu, X.~Hong, and F.~Zhang, ``Efficient and probabilistic
  adaptive voxel mapping for accurate online lidar odometry,'' \emph{IEEE
  Robotics and Automation Letters}, vol.~7, no.~3, pp. 8518--8525, 2022.

\bibitem{fischler1981ransac}
M.~A. Fischler and R.~C. Bolles, ``Random sample consensus: a paradigm for
  model fitting with applications to image analysis and automated
  cartography,'' \emph{Communications of the ACM}, vol.~24, no.~6, pp.
  381--395, 1981.

\bibitem{ceres-solver}
S.~Agarwal, K.~Mierle, and Others, ``Ceres solver,''
  \url{http://ceres-solver.org}.

\bibitem{KITTI}
A.~Geiger, P.~Lenz, and R.~Urtasun, ``Are we ready for autonomous driving? the
  kitti vision benchmark suite,'' in \emph{Conference on Computer Vision and
  Pattern Recognition (CVPR)}, 2012.

\bibitem{NCLT}
N.~Carlevaris-Bianco, A.~K. Ushani, and R.~M. Eustice, ``University of
  {Michigan} {North} {Campus} long-term vision and lidar dataset,''
  \emph{International Journal of Robotics Research}, vol.~35, no.~9, pp.
  1023--1035, 2015.

\bibitem{complex_urban_dataset}
J.~Jeong, Y.~Cho, Y.-S. Shin, H.~Roh, and A.~Kim, ``Complex urban dataset with
  multi-level sensors from highly diverse urban environments,''
  \emph{International Journal of Robotics Research}, vol.~38, no.~6, pp.
  642--657, 2019.

\bibitem{liliom}
K.~Li, M.~Li, and U.~D. Hanebeck, ``Towards high-performance
  solid-state-lidar-inertial odometry and mapping,'' \emph{IEEE Robotics and
  Automation Letters}, vol.~6, no.~3, pp. 5167--5174, 2021.

\bibitem{kaess2008isam}
M.~Kaess, A.~Ranganathan, and F.~Dellaert, ``isam: Incremental smoothing and
  mapping,'' \emph{IEEE Transactions on Robotics}, vol.~24, no.~6, pp.
  1365--1378, 2008.

\bibitem{kaess2012isam2}
M.~Kaess, H.~Johannsson, R.~Roberts, V.~Ila, J.~J. Leonard, and F.~Dellaert,
  ``isam2: Incremental smoothing and mapping using the bayes tree,'' \emph{The
  International Journal of Robotics Research}, vol.~31, no.~2, pp. 216--235,
  2012.

\end{thebibliography}

\end{document}